\documentclass[14pt]{article}

\usepackage{sbc-template}

\usepackage{graphicx,url}

\usepackage[utf8x]{inputenc}
\usepackage{booktabs}
\usepackage{multirow}
\usepackage{graphicx}
\usepackage{array}
\usepackage{todonotes}

\title{Portuguese Word Embeddings: Evaluating on Word Analogies and Natural Language Tasks}

\author{Nathan S. Hartmann\inst{1}, Erick Fonseca\inst{1}, Christopher D. Shulby\inst{1},\\ Marcos V. Treviso\inst{1}, Jéssica S. Rodrigues\inst{2}, Sandra M. Aluísio\inst{1}}

\address{University of São Paulo, Institute of Mathematics and Computer Sciences
\nextinstitute
  Federal University of São Carlos, Department of Computer Science
\email{\{nathansh,erickrf,sandra\}@icmc.usp.br}
\email{\{chrisshulby,marcosvtreviso,jsc\}@gmail.com}
}

\begin{document}

\maketitle

\begin{abstract}
Word embeddings have been found to provide meaningful representations for words in an efficient way; therefore, they have become common in Natural Language Processing systems. In this paper, we evaluated different word embedding models trained on a large Portuguese corpus, including both Brazilian and European variants. We trained 31 word embedding models using FastText, GloVe, Wang2Vec and Word2Vec. We evaluated them intrinsically on syntactic and semantic analogies and extrinsically on POS tagging and sentence semantic similarity tasks. The obtained results suggest that word analogies are not appropriate for word embedding evaluation; task-specific evaluations appear to be a better option.
\end{abstract}

\section{Introduction}


Natural Language Processing (NLP) applications usually take words as basic input units; therefore, it is important that they be represented in a meaningful way.
In recent years, \emph{word embeddings} have been found to efficiently provide such representations, and consequently, have become common in modern NLP systems. They are vectors of real valued numbers, which represent words in an $n$-dimensional space, learned from large non-annotated corpora and able to capture syntactic, semantic and morphological knowledge.



Different algorithms have been developed to generate embeddings \cite[\emph{inter alia}]{bengio2003neural, collobertetal2011,mikolovetal2013, Ling:2015:naacl, lai2015recurrent}. They can be roughly divided into two families of methods \cite{baronietal2014}: the first is composed of methods that work with a co-occurrence word matrix, such as Latent Semantic Analysis (LSA) \cite{dumais1988using}, Hyperspace Analogue to Language (HAL) \cite{lund1996producing} and Global Vectors (GloVe) \cite{penningtonetal2014}. The second is composed of predictive methods, which try to predict neighboring words given one or more context words, such as Word2Vec \cite{mikolovetal2013}.





Given this variety of word embedding models, methods for evaluating them becomes a topic of interest. \cite{mikolovetal2013} developed a benchmark for embedding evaluation based on a series of analogies. Each analogy is composed of two pairs of words that share some syntactic or semantic relationship, e.g., the names of two countries and their respective capitals, or two verbs in their present and past tense forms. In order to evaluate an embedding model, applying some vectorial algebra operation to the vectors of three of the words should yield the vector of the fourth one. A version of this dataset translated and adapted to Portuguese was created by \cite{rodriguesetal2016}. 

However, in spite of being popular and computationally cheap, \cite{repeval:16} suggests that word analogies are not appropriate for evaluating embeddings. Instead, they suggest using task-specific evaluations, i.e., to compare word embedding models on how well they perform on downstream NLP tasks.

In this paper, we evaluated different word embedding models trained on a large Portuguese corpus, including both Brazilian and European variants (Section 2). 
We trained our models using four different algorithms with varying dimensions (Section 3). We evaluated them on the aforementioned analogies as well as on POS tagging and sentence similarity, to assess both syntactic and semantic properties of the word embeddings (Section 4). Section 5 revises recent studies evaluating Portuguese word embeddings and compares literature results with ours. The contributions of this paper are: i) to make a set of 31 word embedding models publicly available\footnote{Available at \url{http://nilc.icmc.usp.br/embeddings}} as well as the script used for corpus preprocessing; and ii) an intrinsic and extrinsic evaluation of word embedding models, indicating the lack of correlation between performance in syntactic and semantic analogies and syntactic and semantic NLP tasks.


\section{Training Corpus}

We collected a large corpus from several sources in order to obtain a multi-genre corpus, representative of the Portuguese language. We rely on the results found by \cite{rodriguesetal2016} and \cite{Fonseca2016} which indicate that the bigger a corpus is, the better the embeddings obtained, even if it is mixed with Brazilian and European texts. Table \ref{tab:corpusembeddings} presents all corpora collected in this work.

\subsection{Preprocessing}

We tokenized and normalized our corpus in order to reduce the vocabulary size, under the premise that vocabulary reduction provides more representative vectors. Word types with less than five occurrences were replaced by a special \texttt{UNKNOWN} symbol. 
Numerals were normalized to zeros;  URL's were mapped to a token \texttt{URL} and emails were mapped to a token \texttt{EMAIL}. 

Then, we tokenized the text relying on whitespaces and punctuation signs, paying special attention to hyphenation. Clitic pronouns like ``machucou-se'' are kept intact. Since it differs from the approach used in \cite{rodriguesetal2016} and their corpus is a subset of ours, we adapted their tokenization using our criteria. We also removed their Wikipedia section, and in all our subcorpora, we only used sentences with 5 or more tokens in order to reduce noisy content. This reduced the number of tokens of LX-Corpus from 1,723,693,241 to 714,286,638.

\begin{table}[!ht]
  \centering
    \scriptsize
    \scalebox{.9}{
    \begin{tabular}{m{2.7cm}llm{2cm}m{7cm}}
      \toprule
        \textbf{Corpus} & \textbf{Tokens} & \textbf{Types} & \textbf{Genre} & \textbf{Description}\\
        \midrule
        LX-Corpus \cite{rodriguesetal2016} & 714,286,638 & 2,605,393 & Mixed genres & A huge collection of texts from 19 sources. Most of them are written in European Portuguese. \\
        \midrule
        Wikipedia & 219,293,003 & 1,758,191 &  Encyclopedic & Wikipedia dump of 10/20/16 \\
        \midrule
        GoogleNews & 160,396,456 & 664,320 & Informative &  News crawled from GoogleNews service\\
        \midrule
        SubIMDB-PT & 129,975,149 & 500,302 & Spoken language & Subtitles crawled from IMDb website\\
        \midrule
        G1 & 105,341,070 & 392,635 & Informative & News crawled from G1 news portal between 2014 and 2015.\\
        \midrule
        PLN-Br \cite{bruckschenetal2008} & 31,196,395 & 259,762 & Informative & Large corpus of the PLN-BR Project with texts sampled from  1994 to 2005. It was also used by \cite{hartmann2016} to train word embeddings models\\
        \midrule
        Literacy works of\newline public domain & 23,750,521 & 381,697 & Prose & A collection of 138,268 literary works from the Domínio Público website \\
        \midrule
        Lacio-web \cite{aluisio2003lacioweb} & 8,962,718 & 196,077 & Mixed genres & Texts from various genres, e.g., literary and its subdivisions (prose, poetry and drama), informative, scientific, law, didactic technical\\
        \midrule
        Portuguese e-books & 1,299,008 & 66,706 & Prose & Collection of classical fiction books written in Brazilian Portuguese crawled from Literatura Brasileira website\\
        \midrule
        Mundo Estranho & 1,047,108 & 55,000 & Informative & Texts crawled from Mundo Estranho magazine\\
        \midrule
        CHC & 941,032 & 36,522 & Informative & Texts crawled from Ci\^encia Hoje das Crian\c{c}as (CHC) website\\
        \midrule
        FAPESP & 499,008 & 31,746 & Science \newline Communication & Brazilian science divulgation texts from Pesquisa FAPESP magazine\\
        \midrule
        Textbooks & 96,209 & 11,597 & Didactic & Texts for children between 3rd and 7th-grade years of elementary school\\
        \midrule
        Folhinha & 73,575 & 9,207 & Informative & News written for children, crawled in 2015 from Folhinha issue of Folha de São Paulo newspaper\\
        \midrule
        NILC subcorpus & 32,868 & 4,064 & Informative & Texts written for children of 3rd and 4th-years of elementary school  \\
        \midrule
        Para Seu Filho Ler & 21,224 & 3,942 & Informative & News written for children, from Zero Hora newspaper \\
        \midrule
        SARESP & 13,308 & 3,293 &  Didactic & Text questions of Mathematics, Human Sciences, Nature Sciences and essay writing to evaluate students\\
        
        \bottomrule
        \\
        \cmidrule{1-3}
        \textbf{Total} & 1,395,926,282 & 3,827,725\\
        \cmidrule{1-3}
    \end{tabular}}
    \caption{Sources and statistics of corpora collected.}
    \label{tab:corpusembeddings}
\end{table}


\section{Embedding Methods}

In this section, we describe the four methods we used to train 31 word embedding models: GloVe, Word2Vec, Wang2Vec, and FastText.



The Global Vectors (GloVe) method was proposed by \cite{penningtonetal2014}, and obtained state-of-the-art results for \emph{syntactic} and \emph{semantic} analogies tasks. This method consists in a co-occurrence matrix $M$ that is constructed by looking at context words. Each element $M_{ij}$ in the matrix represents the probability of the word $i$ being close to the word $j$. In the matrix $M$, the rows (or vectors) are randomly generated and trained by obeying the equation $P(w_i, w_j) = log(M_{ij}) = w_iw_j + b_i + b_j$
, where $w_i$ and $w_j$ are word vectors, and $b_i$ and $b_j$ are biases.





Word2Vec is a widely used method in NLP for generating word embeddings. It has two different training strategies: (i) \emph{Continuous Bag-of-Words (CBOW)}, in which the model is given a sequence of words without the middle one, and attempts to predict this omitted word; (ii) \emph{Skip-Gram}, in which the model is given a word and attempts to predict its neighboring words. In both cases, the model consists of only a single weight matrix (apart from the word embeddings), which results in a fast log-linear training that is able to capture \emph{semantic} information~\cite{mikolovetal2013}. 





Wang2Vec is a modification of Word2Vec made in order to take into account the lack of word order in the original architecture. Two simple modifications were proposed in Wang2Vec expecting embeddings to better capture \emph{syntactic} behavior of words \cite{Ling:2015:naacl}. In the \emph{Continuous Window} architecture, the input is the concatenation of the context word embeddings in the order they occur. In \emph{Structured Skip-Gram}, a different set of parameters is used to predict each context word, depending on its position relative to the target word. 



FastText is a recently developed method \cite{bojanowski2016enriching,joulin2016bag} in which embeddings are associated to character n-grams, and words are represented as the summation of these representations. In this method, a word representation is induced by summing character n-gram vectors with vectors of surrounding words. Therefore, this method attempts to capture \emph{morphological} information to induce word embeddings.


\section{Evaluation}

In order to evaluate the robustness of the word embedding models we trained, we performed intrinsic and extrinsic evaluations. For the intrinsic evaluation, we used the set of syntactic and semantic analogies from \cite{rodriguesetal2016}. For extrinsic evaluation, we chose to apply the trained models on POS tagging and sentence similarity tasks. The tasks were chosen deliberately since they are linguistically aligned with the sets of analogies used in the first evaluation. POS tagging is by nature a morphosyntactic task, and although some analogies are traditionally regarded as \emph{syntactic}, they are actually morphological --- for example, suffix operations. Sentence similarity is a semantic task since it evaluates if two sentences have similar meaning. It is expected that the models that achieve the best results in syntactic (morphological) analogies also do so in POS tagging, and the same is true for semantic analogies and semantic similarity evaluation. We trained embeddings with the following dimensions numbers: 50, 100, 300, 600 and 1,000.


\subsection{Intrinsic evaluation}

We evaluated our embeddings in the syntactic and semantic analogies provided by \cite{rodriguesetal2016}. Since our corpus is composed of both Brazilian (PT-BR) and European (PT-EU) Portuguese, we also evaluated the models in the test sets for both variants, following \cite{rodriguesetal2016}.

Table \ref{tab:evaluation} shows the obtained results for the intrinsic evaluation. On average, GloVe was the best model for both Portuguese variants. The model which best performed on syntactic analogies was FastText, followed by Wang2Vec. This makes sense since FastText is a morphological model, and Wang2Vec uses word order, which provides some minimal syntactic knowledge. In semantic analogies, the model which best performed was GloVe, followed by Wang2Vec. GloVe is known for modeling semantic information well. Wang2Vec potentially captures semantics because it uses word order. The position of a negation in a sentence can totally change its semantics. If this negation is shuffled in a bag of words (Word2Vec CBOW), sentence semantic is diluted.

All CBOW models, except for the Wang2Vec ones, achieved very low results in semantic analogies, similarly to the results from \cite{mikolovetal2013}. 
Wang2Vec CBOW differs from other CBOW methods in that it takes word order into account, and then we can speculate that an unordered bag-of-words is not able to capture a word's semantic so well.


\begin{table}[t]
	\center
    \footnotesize
    \scalebox{.73}{
    \begin{tabular}{llrccc|ccc}
    	\toprule
        \multicolumn{2}{c}{\multirow{2}{*}{\textbf{Embedding Models}}} & \multirow{2}{*}{\textbf{Size}} & \multicolumn{3}{c}{\textbf{PT-BR}} & \multicolumn{3}{c}{\textbf{PT-EU}}\\
        \cmidrule{4-9}
        \multicolumn{2}{c}{} & & \textbf{Syntactic} & \textbf{Semantic} & \textbf{All} & \textbf{Syntactic} & \textbf{Semantic} & \textbf{All}\\
        \midrule
        & & 50 	                & 35.2 & 4.2 & 19.6 & 35.2 & 4.6 & 19.8 \\
        & & 100                 & 45.0 & 6.1 & 25.5 & 45.1 & 6.4 & 25.7  \\
        & CBOW & 300            & 52.0 & 8.4 & 30.1 & 52.0 & 9.1 & 30.5  \\
        \multirow{2}{*}{FastText} & & 600                 & 52.6 & 5.9 & 29.2 & 52.4 & 6.5 & 29.4\\
        & & 1,000 & 50.6 & 4.8 & 27.7 & 50.4 & 5.4 & 27.9\\
        \cmidrule{2-9}
        &  & 50                 & 36.8 & 18.4 & 27.6 & 36.5 & 17.1 & 26.8  \\
        &  & 100 & 50.8 & 30.0 & 40.4 & 50.7 & 28.9 & 39.8   \\
        & Skip-Gram & 300 & \textbf{58.7} & 32.2 & 45.4 & \textbf{58.5} & 31.1 & 44.8  \\
        & & 600 & 55.1 & 24.3 & 39.6 & 55.0 & 23.9 & 39.4\\
        & & 1,000 & 45.1 & 14.6 & 29.8 & 45.2 & 13.8 & 29.4\\
        \midrule         
        &  & 50 & 28.7 & 13.7 & 27.4 & 28.5 & 12.8 & 27.7\\
         & & 100 & 39.7 & 28.7 & 34.2 & 39.9 & 26.6 & 33.2\\
        GloVe & & 300 & 45.8 & 45.8 & \textbf{46.7} & 45.9 & 42.3 & \textbf{46.2}\\
        & & 600 & 42.3 & \textbf{48.5} & 45.4 & 42.3 & \textbf{43.8} & 43.1\\
        & & 1,000 & 39.4 & 45.9 & 42.7 & 39.8 & 42.5 & 41.1\\
        \midrule
		 & & 50 & 28.4 & 9.2 & 18.8 & 28.4 & 8.9 & 18.6\\
        &  & 100 & 40.9 & 26.2 & 33.5 & 40.8 & 24.4 & 32.6\\
         & CBOW & 300 & 49.9 & 40.3 & 45.1 & 50.0 & 36.9 & 43.5\\
        & & 600 & 46.1 & 22.2 & 34.1 & 46.0 & 21.1 & 33.5\\
        \multirow{2}{*}{Wang2Vec} & & 1,000 & 44.8 & 21.9 & 33.3 & 44.7 & 20.5 & 32.6\\
        \cmidrule{2-9}
        &  & 50 & 30.6 & 12.2 & 21.3 & 30.6 & 11.5 & 21.0\\
        &  & 100 & 43.9 & 22.2 & 33.0 & 44.0 & 21.2 & 32.6 \\
        & Skip-Gram& 300 & 53.3 & 33.9 &  42.8 & 53.4 & 32.3 & 43.6\\
        & & 600 & 52.9 & 35.0 & 43.9 & 53.0 & 33.2 & 43.1\\
        & & 1,000 & 47.3 & 33.2 & 40.2 & 47.6 & 30.9 & 39.2\\
        \midrule
        & & 50 & 9.8 & 2.2 & 6.0 & 9.7 & 1.9 & 5.8\\
        & & 100 & 16.2 & 3.6 & 9.9 & 16.0 & 3.5 & 9.7\\
        & CBOW & 300 & 24.7 & 4.6 & 23.9 & 24.5 & 4.5 & 23.6\\
        & & 600 & 25.8 & 5.2 & 23.1 & 25.4 & 5.1 & 22.9\\
        \multirow{2}{*}{Word2Vec} & & 1,000 & 26.2 & 4.9 & 22.9 & 26.2 & 4.5 & 22.7\\
        \cmidrule{2-9}
        &  & 50 & 17.0 & 5.4 & 11.2 & 16.9 & 4.8 & 10.8\\
        & & 100 & 25.2 & 8.0 & 16.6 & 24.8 & 7.4 & 16.1 \\
        & Skip-Gram & 300 & 33.0 & 15.6 & 29.2 & 32.2 & 14.1 & 29.8\\
        & & 600 & 35.6 & 20.0 & 33.4 & 35.3 & 17.6 & 33.5\\
        & & 1,000 & 34.1 & 21.3 & 32.6 & 33.6 & 18.1 & 31.9\\
        \bottomrule
    \end{tabular}}
    \caption{Intrinsic evaluation on syntactic and semantic analogies.}
    \label{tab:evaluation}
\end{table}

\subsection{Extrinsic Evaluation}

In this section we describe the experiments performed on POS tagging and Semantic Similarity tasks.

\subsubsection*{POS Tagging}

  POS tagging is a very suitable NLP task to evaluate how well the embeddings capture morphosyntactic properties. The two key difficulties here are: i) correctly classifying words that can have different tags depending on context; and ii) generalizing to previously unseen words.
    Our experiments were performed with the nlpnet POS tagger\footnote{More info at \url{http://nilc.icmc.usp.br/nlpnet/}} using the revised Mac-Morpho corpus and similar tagger configurations to those presented by \cite{Fonseca2015} (20 epochs, 100 hidden neurons, learning rate starting at 0.01, capitalization, suffix and prefix features). We did not focus on optimizing hyperparameters; instead, we set a single configuration to compare embeddings.
    
    Table~\ref{tab:pos} presents the POS accuracy results\footnote{Note that accuracies are well below those reported by \cite{Fonseca2015}. The probable cause is that the embedding vocabularies used here did not have clitic pronouns split from verbs, resulting in a great amount of out of vocabulary words.}.  As a rule of thumb, the larger the dimensionality, the better the performance. The exception is the 1,000 dimensions Word2Vec models, which performed slightly worse than those with 600. GloVe and FastText yielded the worst results, and Wang2Vec achieved the best. GloVe's poor performance may be explained by its focus on semantics rather than syntax, and FastText's performance was surprising in that despite its preference for morphology, something traditionally regarded as important for POS tagging, it yielded relatively poor results. Wang2Vec resulted in the best performance -- actually, its 300 dimension Skip-Gram model was superior to Word2Vec's 1000 model. Concerning the CBOW and Skip-Gram strategies, in the case of FastText, the latter was considerably better. For Wang2Vec and Word2Vec, the gap between the two is less noticeable, where CBOW achieved slightly better performance on smaller dimensionalities. 

\begin{table}[htb]
\centering
\footnotesize
\scalebox{.73}{
\begin{tabular}[t]{@{}llrr@{}}
\toprule
\multicolumn{2}{c}{\textbf{Embedding Models}}          & \multicolumn{1}{c}{\textbf{Size}} & \multicolumn{1}{c}{\textbf{Accuracy}} \\ \midrule
\multirow{10}{*}{FastText} & \multirow{5}{*}{CBOW}      & 50                                & 91.18\%                               \\
                          &                            & 100                               & 92.57\%                               \\
                          &                            & 300                               & 93.86\%                               \\
                          &                            & 600                               & 93.86\%                               \\ 
                          & & 1000 & 94.27\% \\
                          \cmidrule(l){2-4} 
                          & \multirow{5}{*}{Skip-Gram} & 50                                & 93.15\%                               \\
                          &                            & 100                               & 93.78\%                               \\
                          &                            & 300                               & 94.82\%                               \\
                          &                            & 600                               & 95.25\%                               \\
                          & & 1000 & 95.49\% \\ \midrule
 & \multirow{3}{*}{CBOW}      & 50                                & 95.33\%                               \\
                          &                            & 100                               & 95.59\%                               \\
\multirow{2}{*}{Wang2Vec}                          &                            & 300                               & 95.83\%                               \\ \cmidrule(l){2-4} 
                          & \multirow{5}{*}{Skip-Gram} & 50                                & 95.07\%                               \\
                          &                            & 100                               & 95.57\%                               \\
                          &                            & 300                               & 95.89\%                               \\
 &  & 600                               &       95.88\%                         \\
 & & 1,000 & \textbf{95.94\%} \\
\bottomrule
\end{tabular}}
\quad
\scalebox{0.73}{
\begin{tabular}[t]{@{}llrr@{}}
\toprule
\multicolumn{2}{c}{\textbf{Embeddings model}}           & \multicolumn{1}{c}{\textbf{Size}} & \multicolumn{1}{c}{\textbf{Accuracy}} \\ \midrule
\multirow{5}{*}{GloVe}     & \multirow{4}{*}{}         & 50                                & 93.13\%                               \\
                           &                            & 100                               & 93.72\%                               \\
                           &                            & 300                               & 94.76\%                               \\
                           &                            
			& 600                               & 95.23\%\\
            & & 1,000                               & 95.57\%\\
                           \midrule
\multirow{10}{*}{Word2Vec} & \multirow{5}{*}{CBOW}      & 50                                & 95.00\%                               \\
                           &                            & 100                               & 95.27\%                               \\
                           &                            & 300                               & 95.58\%                               \\
                           &                            & 600                               & 95.65\%                               \\
                           &                            & 1,000                              & 95.62\%                               \\ \cmidrule(l){2-4} 
                           & \multirow{5}{*}{Skip-Gram} & 50                                & 94.79\%                               \\
                           &                            & 100                               & 95.18\%                               \\
                           &                            & 300                               & 95.66\%                               \\
                           &                            & 600                               & 95.82\%                               \\
                           &                            & 1,000                              & 95.81\%                               \\ \bottomrule
\end{tabular}}
\caption{Extrinsic evaluation on POS tagging}
\label{tab:pos}
\end{table}

\subsubsection*{Semantic Similarity}

ASSIN (\textit{Avaliação de Similaridade Semântica e Inferência Textual}) was a workshop co-located with PROPOR-2016. ASSIN made two shared-tasks available: i) semantic similarity; and ii) entailment. We chose the first one to evaluate our word embedding models extrinsically in a semantic task. ASSIN semantic similarity shared task required participants to assign similarity values between 1 and 5 to pairs of sentences. The workshop made training and test sets for Brazilian (PT-BR) and European (PT-EU) Portuguese available. \cite{hartmann2016} obtained the best results for this task. The author calculated the semantic similarity of pairs of sentences training a linear regressor with two features: i) the cosine similarity between the TF-IDF of each sentence; and ii) the cosine similarity between the summation of the word embeddings of the sentences' words. We chose this work as a baseline for evaluation because we can replace its word embedding model with others and compare the results. Although the combination of TF-IDF and word embeddings produced better results than only using word embeddings, we chose to only use embeddings for ease of comparison. \cite{hartmann2016} trained the word embedding model using Word2Vec Skip-Gram approach, with 600 dimensions, and a corpus composed of Wikipedia, G1 and PLN-Br. Only using embeddings, \cite{hartmann2016} achieved 0.58 in Pearson's Correlation ($\rho$) and a 0.50 Mean Squared Error (MSE) for PT-BR; and 0.55 $\rho$ and 0.83 MSE for PT-EU evaluation.

Table \ref{tab:semantic_similarity_evaluation} shows the performance of our word embedding models for both PT-BR and PT-EU test sets. To our surprise, the word embedding models which achieved the best results on semantic analogies (see Table \ref{tab:evaluation}) were not the best in this semantic task. The best results for European Portuguese was achieved by Word2Vec CBOW model using 1,000 dimensions. CBOW models were the worst on semantic analogies and were not expected to achieve the best results in this task. The best result for Brazilian Portuguese was obtained by Wang2Vec Skip-Gram model using 1,000 dimensions. This model also achieved the best results for POS tagging. Neither FastText nor GloVe models beat the results achieved by \cite{hartmann2016}.

\begin{table}[htb]
\centering
\footnotesize
\scalebox{0.73}{
\begin{tabular}[t]{llrrr|rr}
\toprule
\multicolumn{2}{c}{\multirow{2}{*}{\textbf{Embedding Models}}} & \multirow{2}{*}{\textbf{Size}} & \multicolumn{2}{c}{\textbf{PT-BR}} & \multicolumn{2}{c}{\textbf{PT-EU}} \\
\cmidrule{4-7}
 \multicolumn{2}{c}{} & & \textbf{$\rho$} & \textbf{MSE} & \textbf{$\rho$} & \textbf{MSE}\\
\midrule

&  & 50 & 0.36 & 0.66 & 0.34 & 1.05\\
& & 100 & 0.37 & 0.66 & 0.36 & 1.04\\
& CBOW & 300 & 0.38 & 0.65 & 0.37 & 1.03\\
\multirow{2}{*}{FastText} & & 600 & 0.33 & 0.68 & 0.38 & 1.02\\
& & 1,000 & 0.39 & 0.64 & 0.41 & 0.99\\
\cmidrule{2-7}
& & 50 & 0.45 & 0.61 & 0.43 & 0.98\\
& & 100 & 0.49 & 0.58 & 0.47 & 0.94\\
& Skip-Gram & 300 & 0.55 & 0.53 & 0.40 & 1.02\\
& & 600 & 0.40 & 0.64 & 0.40 & 1.01\\
& & 1,000 & 0.52 & 0.56 & 0.54 & 0.86\\
\midrule
 &  & 50 & 0.53 & 0.55 & 0.51 & 0.89\\
& & 100 & 0.56 & 0.52 & 0.54 & 0.85\\
& CBOW & 300  & 0.53 & 0.55 & 0.51 & 0.89\\
& & 600 & 0.49 & 0.58 & 0.53 & 0.87\\
\multirow{2}{*}{Wang2Vec}& & 1,000 & 0.50 & 0.57 & 0.53 & 0.87\\
\cmidrule{2-7}
& & 50 & 0.51 & 0.56 & 0.47 & 0.92\\
&  & 100 & 0.54 & 0.54 & 0.50 & 0.89\\
& Skip-Gram & 300 & 0.58 & 0.50 & 0.53 & 0.85\\
& & 600 & 0.59 & \textbf{0.49} & 0.54 & \textbf{0.83}\\
& & 1,000 & \textbf{0.60} & \textbf{0.49} & 0.54 & 0.85\\

\bottomrule
\end{tabular}}
\quad
\scalebox{0.73}{
\begin{tabular}[t]{llrrr|rr}
\toprule
\multicolumn{2}{c}{\multirow{2}{*}{\textbf{Embedding Models}}} & \multirow{2}{*}{\textbf{Size}} & \multicolumn{2}{c}{\textbf{PT-BR}} & \multicolumn{2}{c}{\textbf{PT-EU}} \\
\cmidrule{4-7}
 \multicolumn{2}{c}{} & & \textbf{$\rho$} & \textbf{MSE} & \textbf{$\rho$} & \textbf{MSE}\\
\midrule
& & 50 & 0.42 & 0.62 & 0.38 & 1.01\\
& & 100 & 0.45 & 0.60 & 0.42 & 0.98\\
GloVe & & 300 & 0.49 & 0.58 & 0.45 & 0.95\\
& & 600 & 0.50 & 0.57 & 0.45 & 0.94\\
& & 1,000 & 0.51 & 0.56 & 0.46 & 0.94\\
\midrule
&  & 50 & 0.47 & 0.59 & 0.46 & 0.95\\
&  & 100 & 0.50 & 0.57 & 0.49 & 0.91\\
& CBOW & 300 & 0.55 & 0.53 & 0.54 & 0.87\\
& & 600 & 0.57 & 0.51 & \textbf{0.55} & 0.86\\
\multirow{2}{*}{Word2Vec} & & 1,000 & 0.58 & 0.50 & \textbf{0.55} & 0.86\\
\cmidrule{2-7}
& & 50 & 0.46 & 0.60 & 0.43 & 0.97\\
& & 100 & 0.48 & 0.58 & 0.45 & 0.95\\
& Skip-Gram & 300 & 0.52 & 0.56 & 0.48 & 0.93\\
& & 600 & 0.53 & 0.54 & 0.50 & 0.92\\
& & 1,000 & 0.54 & 0.54 & 0.50 & 0.91\\
\bottomrule
\end{tabular}}
\caption{Extrinsic evaluation on Semantic Similarity task.}
\label{tab:semantic_similarity_evaluation}
\end{table}
\section{Related Work}
  
The research on evaluating unsupervised word embeddings can be divided into intrinsic and extrinsic evaluations. The former relying mostly on word analogies (e.g. \cite{mikolovetal2013}) and measuring the semantic similarity between words (e.g. the WS-353 dataset \cite{finkelstein2001}), while extrinsic evaluations focus on practical NLP tasks  (e.g. \cite{2016nayak-veceval}). POS tagging, parsing, semantic similarity between sentences, and sentiment analysis are some commonly used tasks for this end.

To the best of our knowledge, only a few works attempted to evaluate Portuguese word embeddings.
\cite{rodriguesetal2016} collected a corpus of Portuguese texts to train word embedding models using the Skip-Gram Word2Vec technique. The authors also translated the benchmark of word analogies developed by \cite{mikolovetal2013} and made it available\footnote{\url{https://github.com/nlx-group/lx-dsemvectors}} for both Brazilian and European Portuguese. The benchmark contains five types of semantic analogy: (i) common capitals and countries, (ii) all capitals and countries, (iii) currency and countries, (iv) cities and states, and (v) family relations. Moreover, nine types of syntactic analogy are also represented: adjectives and adverbs, opposite adjectives, base adjectives and comparatives, base adjectives and superlatives, verb infinitives and present participles, countries and nationalities (adjectives), verb infinitives and past tense forms, nouns in plural and singular, and verbs in plural and singular. They report a 52.8\% evaluation accuracy of their word embedding model in both syntactic and semantic analogies.


\cite{sousa2016} investigated whether Word2Vec (CBOW and Skip-Gram) or GloVe performed best on the benchmark in \cite{rodriguesetal2016}. The author compiled a sample of texts from Wikipedia in Portuguese, searching for articles related to teaching, education, academics, and institutions. The best results were obtained using Word2Vec CBOW to train vectors of 300 dimensions. This model achieved an accuracy of 21.7\% on syntactic analogies, 17.2\% on semantic analogies and 20.4\% overall.

\cite{Fonseca2015} compared the performance of three different vector space models used for POS tagging with a neural tagger. They used Word2Vec Skip-Gram, HAL, and the neural method from \cite{collobertetal2011}; Skip-Gram obtained the best results in all tests.

Concerning the differences between embeddings obtained from Brazilian and European Portuguese texts, \cite{Fonseca2016} present an extrinsic analysis on POS tagging. They trained different embedding models; one with only Brazilian texts, one with only European ones and another with mixed variants; and trained neural POS taggers which were evaluated on Brazilian and European datasets. One of their findings is that, as a rule of thumb, the bigger the corpus in which embeddings are obtained, the better. Additionally, mixing both variants in the embedding generation did not decrease tagger performance in any of the POS test sets. This supports the hypothesis that a single, large corpus comprising Brazilian and European texts can be useful for most NLP applications in Portuguese.
\section{Conclusions and Future Work}

In this paper, we presented the word embeddings we trained using four different techniques and their evaluation.
All trained models are available for download, as well as the script used for corpus preprocessing. The results obtained from intrinsic and extrinsic evaluations were not aligned with each other, contrary to the expected. GloVe produced the best results for syntactic and semantic analogies, and the worst, together with FastText, for both POS tagging and sentence similarity. These results are aligned with those from \cite{repeval:16}, which suggest that word analogies are not appropriate for evaluating word embeddings. Overall, Wang2Vec vectors yielded very good performance across our evaluations, suggesting they can be useful for a variety of NLP tasks.
As future work, we intend to try different tokenization and normalization patterns, and also to lemmatize certain word categories like verbs, since this could significantly reduce vocabulary, allowing for more efficient processing. An evaluation with more NLP tasks would also be beneficial to our understanding of different model performances.



\section*{Acknowledgements}

This work was supported by CNPq, CPqD and FAPESP (PIPE-PAPESP project nº 2016/00500-1).

\bibliographystyle{sbc}
\bibliography{references}

\end{document}